\documentclass[10pt,twocolumn,letterpaper]{article}

\usepackage{iccv}
\usepackage{times}
\usepackage{epsfig}
\usepackage{graphicx}
\usepackage{amsmath}
\usepackage{amssymb}
\usepackage{amsthm}
\usepackage[font=small,skip=2pt]{caption}
\usepackage[inline]{enumitem}
\usepackage[table, dvipsnames]{xcolor}
\usepackage{graphicx}
\usepackage{makecell}
\usepackage[hang,flushmargin]{footmisc}
\usepackage{multirow, booktabs}
\usepackage{mathtools}
\usepackage{pbox}
\usepackage{tabu}
\usepackage{MnSymbol,wasysym}
\usepackage[tracking=true]{microtype}
\usepackage{adjustbox}

\usepackage[pagebackref=true,breaklinks=true,letterpaper=true,colorlinks,bookmarks=false]{hyperref}
 \iccvfinalcopy 

\def\iccvPaperID{1473} 

\DeclareMathOperator*{\argmin}{argmin} 

\ificcvfinal\fi

\newcommand\narrowstyle{\SetTracking{encoding=*}{-50}\lsstyle}
\newcommand\normalstyle{\SetTracking{encoding=*}{0}\lsstyle}

\usepackage{ifthen}
\usepackage[normalem]{ulem}

\newcommand{\michal}[2][]{%
    \ifthenelse{ \equal{#1}{} }
        {\textcolor{Dandelion}{(MI) #2}}
        {\textcolor{Dandelion}{(MI) \sout{#1} #2}}
}
\newcommand{\niv}[2][]{%
    \ifthenelse{ \equal{#1}{} }
        {\textcolor{Orchid}{(NG) #2}}
        {\textcolor{Orchid}{(NG) \sout{#1} #2}}
}
\newcommand{\assaf}[2][]{%
    \ifthenelse{ \equal{#1}{} }
        {\textcolor{Lavender}{(AS) #2}}
        {\textcolor{Lavender}{(AS) \sout{#1} #2}}
}
\newcommand{\ben}[2][]{%
    \ifthenelse{ \equal{#1}{} }
        {\textcolor{JungleGreen}{(BF) #2}}
        {\textcolor{JungleGreen}{(BF) \sout{#1} #2}}
}
\newcommand{\shai}[2][]{%
    \ifthenelse{ \equal{#1}{} }
        {\textcolor{RoyalBlue}{(SB) #2}}
        {\textcolor{RoyalBlue}{(SB) \sout{#1} #2}}
}
\newcommand{\comment}[1]{}

\renewcommand{\michal}[2][]{\ifthenelse{\equal{#1}{}}{}{\textcolor{Dandelion}{#2}}}
\renewcommand{\niv}[2][]{\ifthenelse{\equal{#1}{}}{}{\textcolor{Orchid}{#2}}}
\renewcommand{\assaf}[2][]{\ifthenelse{\equal{#1}{}}{}{\textcolor{Lavender}{#2}}}
\renewcommand{\ben}[2][]{\ifthenelse{\equal{#1}{}}{}{\textcolor{JungleGreen}{#2}}}
\renewcommand{\shai}[2][]{\ifthenelse{\equal{#1}{}}{}{\textcolor{RoyalBlue}{#2}}}

\renewcommand{\michal}[2][]{\ifthenelse{\equal{#1}{}}{}{#2}}
\renewcommand{\niv}[2][]{\ifthenelse{\equal{#1}{}}{}{#2}}
\renewcommand{\assaf}[2][]{\ifthenelse{\equal{#1}{}}{}{#2}}
\renewcommand{\ben}[2][]{\ifthenelse{\equal{#1}{}}{}{#2}}
\renewcommand{\shai}[2][]{\ifthenelse{\equal{#1}{}}{}{#2}}

\definecolor{in_red}{RGB}{252,1,2}
\definecolor{ref_blue}{RGB}{65,116,197}
\definecolor{gan_green}{RGB}{0,153,0}
\definecolor{gpnn_purple}{RGB}{214,74,211}

\begin{document}

\title{\vspace*{-1.4cm}
Drop the GAN:\\ In Defense of Patches Nearest Neighbors as Single Image Generative Models }
\author{Niv Granot$^*$ \qquad
Ben Feinstein$^*$ \qquad
Assaf Shocher$^*$ \qquad
Shai Bagon$^\dagger$ \qquad
Michal Irani$^*$ \\
\small{$^*$Dept. of Computer Science and Applied Math, The Weizmann Institute of Science}
\\
\small{$^\dagger$Weizmann Artificial Intelligence Center (WAIC)}\\
\small
{\textbf{\emph{Project Website:}}}  \url{http://www.wisdom.weizmann.ac.il/\~vision/gpnn/}

}
\twocolumn[{%
\renewcommand\twocolumn[1][]{#1}%
\maketitle

\centering
\vspace{-0.7cm}
\includegraphics[width=\textwidth]{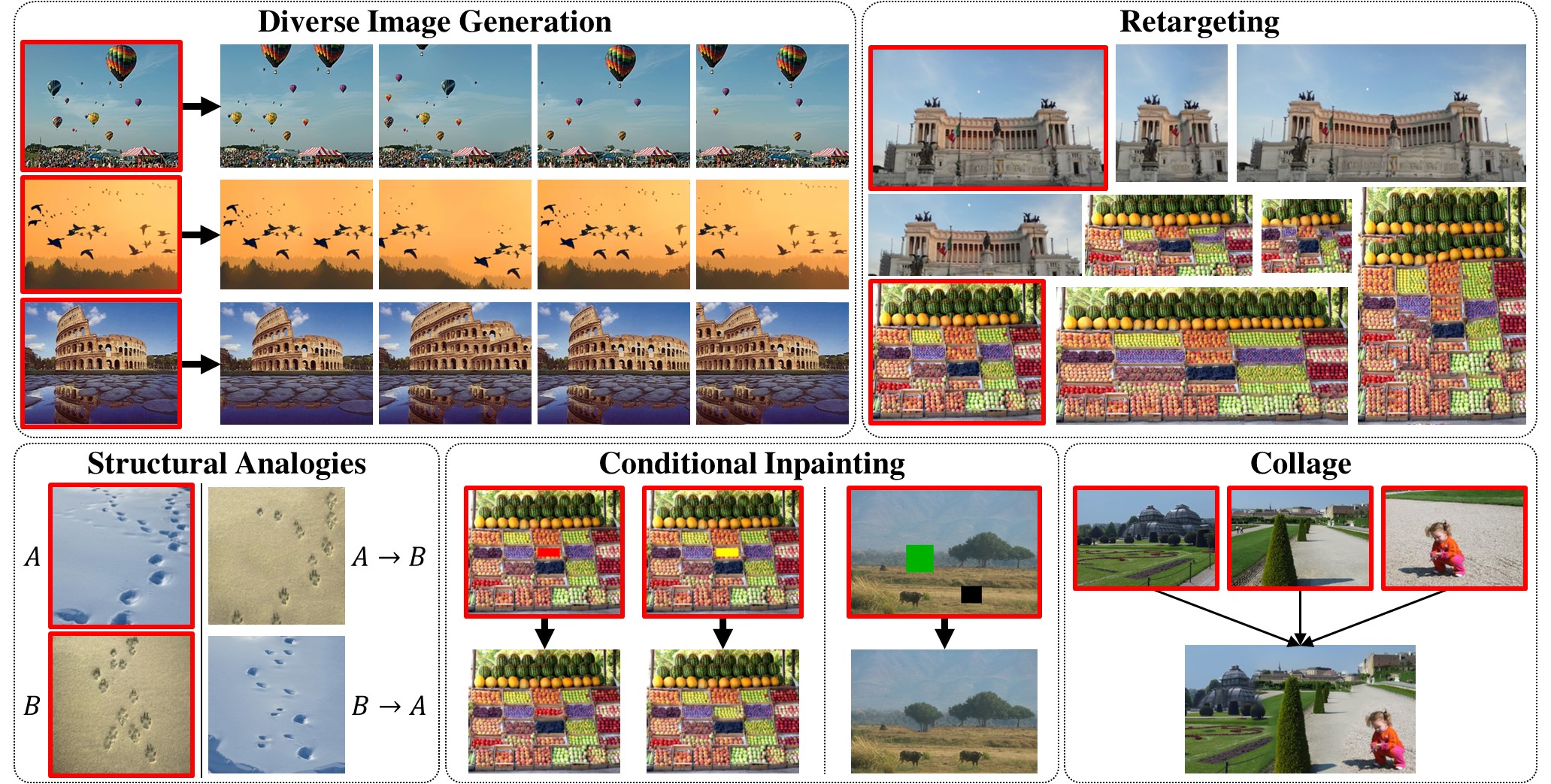}
\captionof{figure}{\small \textit{Our simple unified framework covers a broad spectrum of single-image generative tasks, that usually require hours of training per image for GANs. Using patch nearest neighbors and a single source image, we can perform these tasks in a few seconds and with higher quality. We show here results obtained with our method for pivotal examples shown in SinGAN \cite{shaham2019singan}, InGAN \cite{ingan}, Structural analogies \cite{benaim2020structural} and Bidirectional Similarity \cite{simakov2008summarizing}.  Additionally, we introduce novel applications such as Conditional-Inpainting. Input images marked in red.}
}
\label{fig:teaser}
\vspace{0.4cm}
}]
\ificcvfinal\thispagestyle{empty}\fi

\begin{abstract}
\vspace*{-0.3cm}
%
Single image generative models perform synthesis and manipulation tasks by capturing the distribution of patches within a single image. The classical (pre Deep Learning) prevailing approaches for these tasks are based on an optimization process that maximizes patch similarity between the input and generated output. 
Recently, however, Single Image GANs were introduced both as a superior solution for such manipulation tasks, but also for remarkable novel generative tasks. 
Despite their impressiveness, single image GANs require long training time (usually hours) for each image and each task. They often \ben[produce]{suffer from} artifacts and are prone to optimization issues such as mode collapse.
In this paper, we show that all of these tasks can be performed without any training, within several seconds, in a unified, surprisingly simple framework. We revisit and cast the ``good-old" patch-based methods into a novel optimization-free framework. We start with an initial coarse guess, and then simply
refine
the details coarse-to-fine using patch-nearest-neighbor search. This allows generating random novel images better and much faster than GANs. We further demonstrate a wide range of applications, such as image editing and reshuffling, retargeting to different sizes, structural analogies, image collage and a newly introduced task of conditional inpainting. Not only is our method faster ($\times 10^3$-$\times 10^4$ than a GAN), it produces superior results (confirmed by quantitative and qualitative evaluation), less artifacts and more realistic global structure than any of the previous approaches (whether GAN-based or classical patch-based).
\end{abstract}

\vspace{-0.3cm}
\section{Introduction}

Single-image generative models perform image synthesis and manipulation by capturing the patch distribution of a single image. Prior to the Deep-Learning revolution, the classical prevailing methods 
were based on optimizing the similarity of small patches between the input image and the generated output image. These \emph{unsupervised} patch-based methods (e.g.,~\cite{efros1999texture,simakov2008summarizing,barnes2009patchmatch,pritch2009shiftmap,dekel2015revealing, ren2016examplebased})
gave rise to a wide variety of remarkable image synthesis and manipulation tasks, including image completion, texture synthesis, image summarization/retargeting, collages, image reshuffling, and more. Specifically, the Bidirectional similarity approach~\cite{simakov2008summarizing, barnes2009patchmatch} encourages the output image to contain only patches from the input image (``Visual Coherence''), and vice versa,  the input should contain only patches from the output (``Visual Completeness''). Hence, no new artifacts are introduced in the output image and no critical information is lost either.


Recently, deep \emph{Single-Image Generative Models} took the field of image manipulation by a storm.
These models are a natural extension of ``Deep Internal Learning" \cite{Shocher_2018_CVPR, Gandelsman_2019_CVPR,zhang2019zero,zuckerman2020across,zhang2019internal,ghosh2020depth,ulyanov2018deep}. They train a GAN on a single image, in an unsupervised way, and have shown to produce impressive generative diversity of results, as well as notable new generative tasks. 
Being fully convolutional, these single-image GANs learn the patch distribution of the single input image, and are then able to generate a plethora of new images with the same patch distribution. These include SinGAN~\cite{shaham2019singan} for generating a  \emph{large diversity} of different image instances -- all sampled from the input patch distribution, InGAN~\cite{ingan} for flexible image retargeting, Structural-Analogies~\cite{benaim2020structural}, texture synthesis~\cite{jetchev2016texture,bergmann2017learning,zhou2018non, zhao2021solid}, and more~\cite{hinz2021improved,mastan2020dcil,mastan2021deepcfl,gur2020hierarchical,vinker2020deep,lin2020tuigan, chen2021mogan}.

However, despite their remarkable capabilities (both diverse generated outputs and diverse new tasks), single-image GANs come with several heavy penalties compared to their simpler classical patch-based counterparts:
(i)~They require very long training time (usually hours) for each input image and each task (as opposed to fast patch-based methods~\cite{barnes2009patchmatch}). \ 
(ii)~They are prone to optimization issues such as mode collapse.
(iii)~They often produce poorer visual quality than the classical patch-based methods. 

Hence, while the course of history has taken the field of image synthesis, from the patch search methods to powerful GANs, it turns out that ``good-old'' patch-based methods are superior in many aspects. In this paper we suggest to reconsider simple patch nearest neighbors again, \emph{but with a new twist}, which allows to inject new Single Image GANs-like generative capabilities into  nearest-neighbor patch-based methods, thus obtaining \emph{the best of both worlds}. We further analyze and characterize the pros and cons of these 2 approaches (patch nearest-neighbors vs. single-image GANs) in Sec.~\ref{sec:discussion}. 

We observe that the generative \emph{diversity} of single-image GAN methods 
(which classical methods lack),
stems primarily from their \emph{unconditional} input at coarser image scales. We further observe that the main source of the above-mentioned drawbacks (slowness, mode-collapse, and poorer visual quality compared to classical methods) stems from the generative (GAN-based) module.
We therefore suggest to ``drop the GAN''
\footnote{\noindent
While we could not resist this wordplay, we do acknowledge that single-image GANs have several significant capabilities which cannot be realized by simple patch nearest-neighbors. We discuss these pros (and cons) in Sec.~\ref{sec:discussion}. 
No GANs were harmed in the preparation of this paper... {\large \smiley}
}, 
and replace this module with a simple (upgraded) Patch Nearest-Neighbor (PNN) module,  while maintaining the unconditional nature of GAN-based methods. This gives rise to a simple new \emph{generative} patch-based algorithm, which we call \emph{Generative Patch Nearest-Neighbor} (\textbf{GPNN}). Its noise input \niv{our input is not just noise, I think we can say "Its noise injected input..."} at coarse-levels  
yields \emph{diverse image generation},
of much higher quality and significantly faster than single-image GANs (with similar
diversity). This is verified via extensive evaluations (Sec.~\ref{sec:results}).

GPNN can perform the new generative tasks of single-image GANs, as well as the old classical tasks, in a single unified framework, without any training, in an optimization-free manner, within a few seconds. 
Unlike single-image GAN-based methods, GPNN is very fast 
($\times 10^3$-$\times 10^4$ faster), and produces superior visual results (confirmed by extensive quantitative and qualitative evaluation). Unlike classical patch-based methods, GPNN enjoys the non-deterministic nature of GANs, their large diversity of possible outputs, and new generative tasks/capabilities.  

We demonstrate a wide range of applications of GPNN: first and foremost \emph{diverse image generation}, but also image editing, image retargeting,
structural analogies, image collages, and a newly introduced task of ``conditional inpainting''. We show that 
GPNN produces results which are either comparable or of higher quality
than any of the previous approaches (whether GAN-based or classical patch-based).
We hope  GPNN will serve as a new baseline for single image generation/manipulation tasks.

\begin{figure}[b]
    \vspace*{-0.5cm}
    \centering
    \includegraphics[width=\columnwidth]{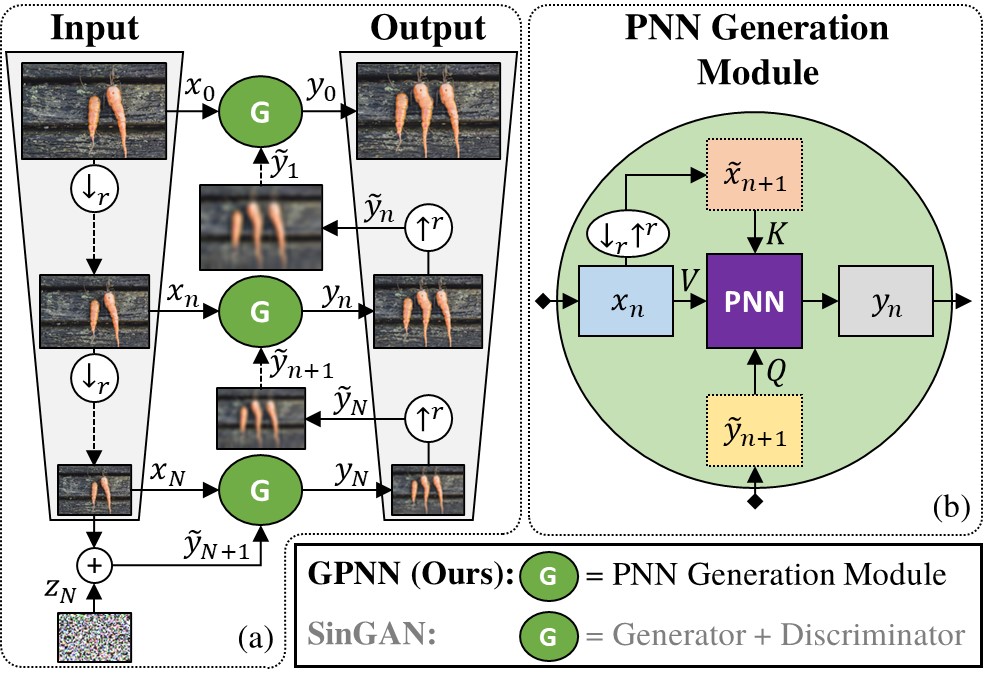}
    \caption{\small \textbf{The GPNN method.}
    \textit{GPNN's multi-scale architecture is very similar to that of SinGAN~\cite{shaham2019singan}: Each scale consists of a single image generator $G$, that generates diverse outputs $y_n$ with similar patch distribution as the source $x_n$. The generation module $G$ (a GAN in~\cite{shaham2019singan}), is replaced here with a non-parametric PNN Generation Module. The coarsest level input is injected with noise.}
    }
    \label{fig:method}
\end{figure}

\vspace{-0.1cm}
\smallskip\noindent Our contributions are therefore several fold:
\begin{itemize}[topsep=0pt,itemsep=-1ex,partopsep=1ex,parsep=1ex,leftmargin=*]
\item We show that ``good old'' patch nearest neighbor approaches can be cast as a generative model, which substantially outperforms modern single-image GANs -- both in quality \shai[(evaluated extensively with many measures)]{} and in speed
\shai[(from hours to seconds)]{}. \niv{Shouldn't we state the measures? Or at least the user study?}
\item We introduce such a casting -- GPNN, a new generative patch-based algorithm, as an alternative to single-image GANs, which provides a unified framework for a large variety of applications.
%
%
\item We analyze and discuss the \emph{inherent pros \& cons} of the modern {GAN-based approaches} 
vs.  classical {Patch-based approaches}.
{We experimentally} characterize the extent to which single-image GANs perform nearest-neighbor extraction behind the scenes. \shai{shouldn't this be a new bullet?}
\end{itemize}

\section{Method}\label{sec:method}

Our goal is to efficiently cast patch nearest neighbor search as a diverse single-image generative model.
To achieve that, GPNN uses a multi-scale architecture with \emph{noise injected input} (Fig.~\ref{fig:method}(a); Sec.~\ref{sec:multi-scale}), similarly to SinGAN~\cite{shaham2019singan}.
However, in each scale, GPNN uses a \emph{non-parametric} Patch Nearest Neighbor (PNN) Generation Module (Fig.~\ref{fig:method}(b), Sec.~\ref{sec:pnn}), as opposed to a full-scale GAN in SinGAN. PNN generates new images with similar patch distribution as the source image \emph{at that scale}.



\subsection{Multi-scale Architecture}\label{sec:multi-scale}
It was previously recognized (e.g.,~\cite{shaham2019singan, simakov2008summarizing, ingan, ren2016examplebased}), that different information is captured in each scale of an image -- from global arrangement at coarser scales, to textures and fine details at finer scales. To capture details from all scales, GPNN has a coarse-to-fine architecture (illustrated in Fig.~\ref{fig:method}(a)). Given a source image $x$, it builds a pyramid $\left\{x_0, \dots, x_N\right\}$, where $x_n$ is $x$ downscaled by a factor $r^n$ (for $r{>}1$; in our current implementation $r{=}\frac{4}{3}$). GPNN uses the same patch size $p \times p$ in all scales ($p{=}7$ in our implementation). Similarly to \cite{shaham2019singan}, the depth of the pyramid $N$ is chosen such that $p$ is approximately half of the image height. At scale $n$, a new image $y_n$ is generated by PNN Generation Module (see Fig.~\ref{fig:method}(b); Sec.~\ref{sec:pnn}), using \emph{real} patches from the source image at that scale $x_n$, guided by $\tilde{y}_{n+1}$, the initial guess. PNN enforces similarity between the internal statistics of the output image and source image at each scale.


\begin{figure}
    \vspace*{-0.3cm}
    \centering
    \includegraphics[width=\columnwidth]{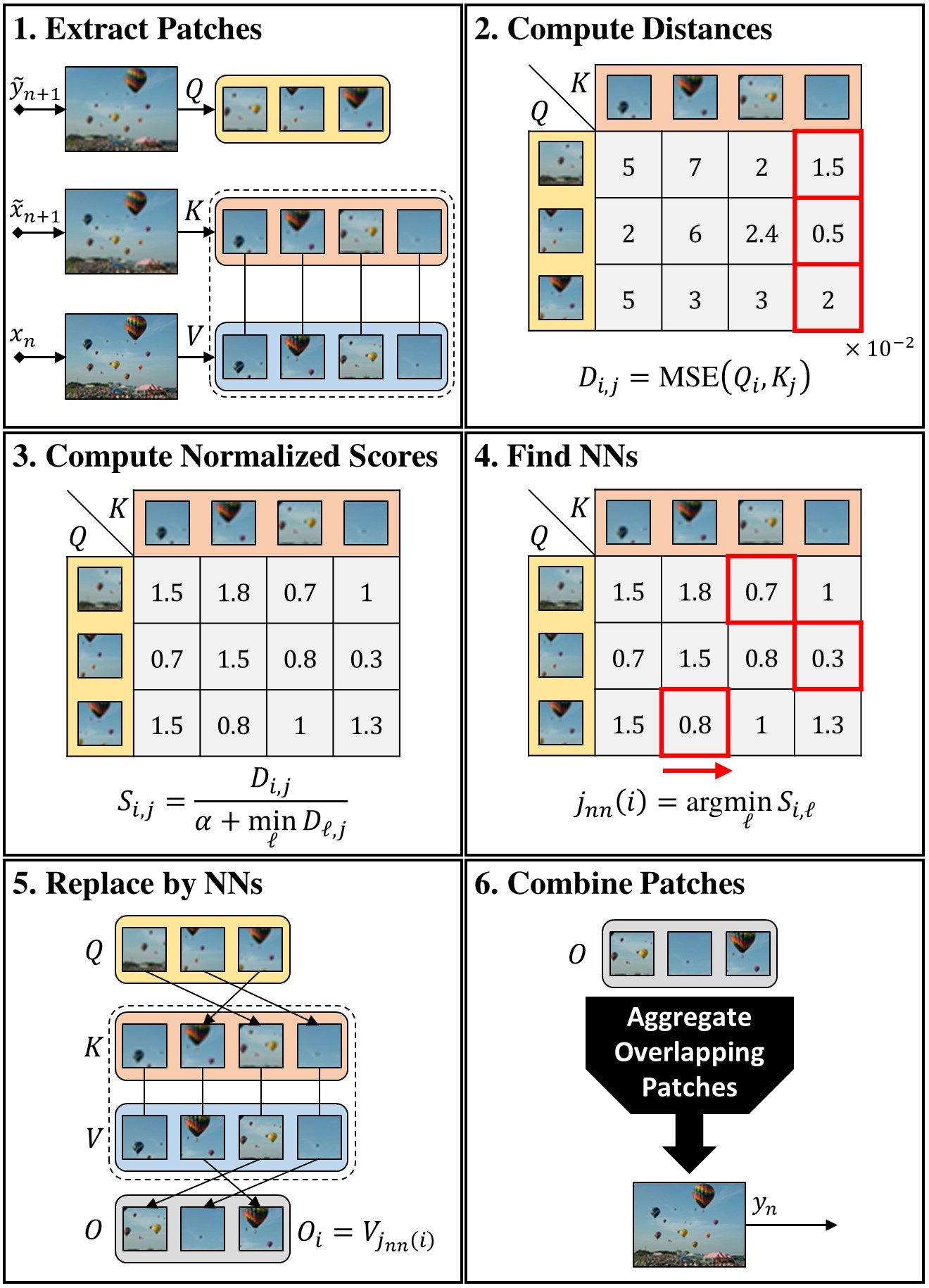}
    \caption{\small \textbf{Algorithmic steps of PNN.}}
    \label{fig:pnn}
    \vspace{-0.5cm}
\end{figure}


In the coarsest level, the initial guess is the source image injected with noise (similarly to \cite{Elad_2017}), $\tilde{y}_{N+1} = x_N + z_N$, where $z_N \sim \mathcal{N}(0,\sigma^2)$ . The coarsest scale defines the arrangement of objects in the image. Injecting noise at that scale makes the nearest-neighbors search nearly random (the mean of the patches remains the same in expectation), hence induces diversity in the global arrangement, yet the PNN maintains coherent outputs. The use of different noise maps $z_N$ is the basis for the diverse image generation presented in Sec.~\ref{sec:results}, whereas different choices for the initial guess are the basis for a wide variety of additional applications we present in Sec.~\ref{sec:apps}.

In finer scales, the initial guess is the upscaled output of the coarser level, $\tilde{y}_{n+1} = y_{n+1} {\uparrow^r}$. The output at each scale is a refinement of the coarser scale output. Hence, the final output $y=y_0$ shares the internal statistics of $x$ at all scales.

\begin{figure*}[t!]
\vspace*{-0.3cm}
\centering
\includegraphics[width=\textwidth]{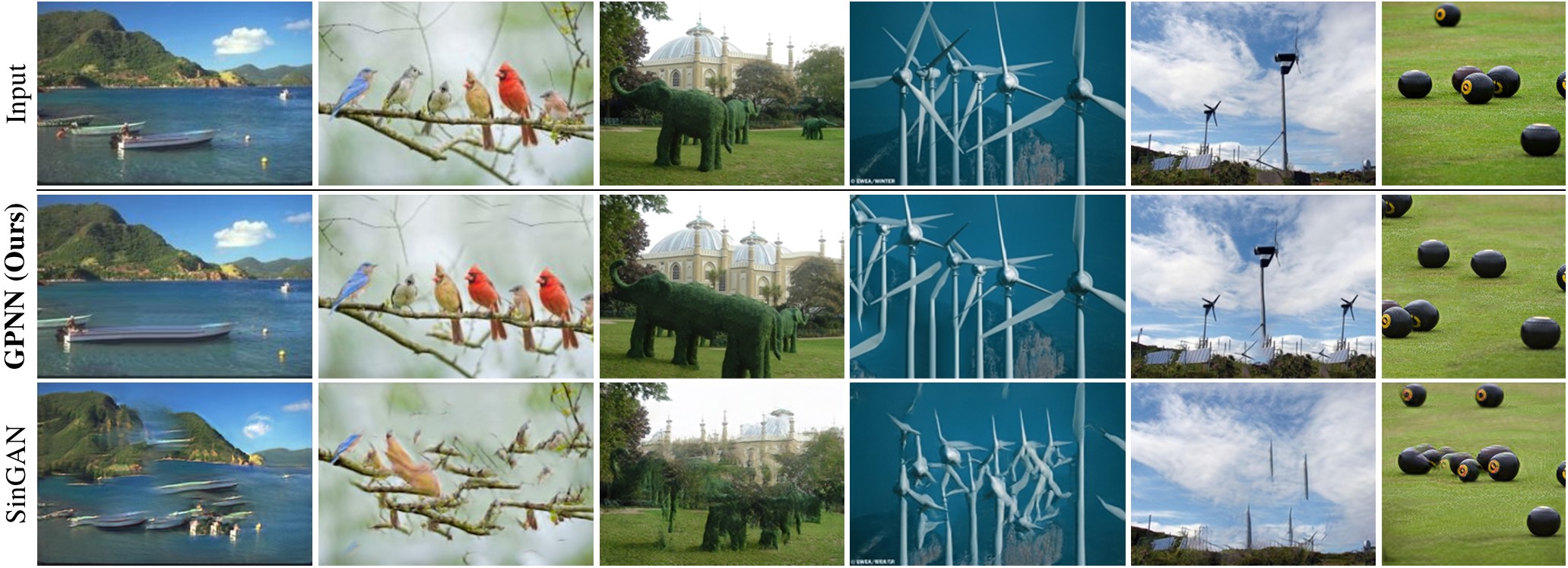}
\caption[]{\small \textbf{Random Instance Generation Comparison:} \textit{(Please zoom in) Images generated by our method are compared with images generated by SinGAN~\cite{shaham2019singan}. 
         Images generated by GPNN (2nd row) look very realistic, whereas SinGAN produces many artifacts (3rd row).}
    }
\label{fig:generation_comparison}
\vspace*{-0.5cm}
\end{figure*}


\subsection{\mbox{Patch Nearest Neighbors Generation Module}}\label{sec:pnn}


The goal of the PNN Generation Module is to generate a new image $y_n$, based on an initial guess image $\tilde{y}_{n+1}$ and a source image $x_n$, such that the structure would be similar to that of $\tilde{y}_{n+1}$'s and the internal statistics would match that of $x_n$. To achieve that, PNN replaces patches from the initial guess $\tilde{y}_{n+1}$ with patches from the source image $x_n$. 

While this coarse-to-fine refinement strategy bears resemblance to that of classical patch-based methods, GPNN introduces 2 major differences (in addition to the unconditional input at coarse scales): \ (i)~A \emph{Query-Key-Value patch search strategy}, which improves the visual quality of the generated output; and \ (ii)~A new \emph{normalized patch-similarity measure}, which ensures visual completeness in an \emph{optimization-free} manner. These differences are detailed below.

Classical patch-based methods use a Query-Reference scheme, where the query is the initial guess and the reference is the source image. Each \emph{query} patch (from the initial guess image) is replaced, or optimized to get closer to, a \emph{reference} patch (from the source image). This encourages similarity between the internal statistics of the output and the source. However, that scheme may fail when there is a significant distribution shift between query and reference patches. For example, when the query patches are blurry (due to upscaling the initial guess from a coarser resolution), they might be matched to blurry reference patches.

To overcome this problem, PNN uses a Query-Key-Value scheme (see Figs.~\ref{fig:method}(b), \ref{fig:pnn}, similarly to \cite{vaswani2017attention}). Instead of comparing the query patch to a \emph{reference} patch and replacing it by that same patch (as done in classical methods), here the lookup patch and replacement patch are different. For each query patch $Q_i$ in $\tilde{y}_{n+1}$, we find its closest \emph{key} patch $K_j$ in a (blurry) upscaled version of the reference image \ben[$x_{n+1}{\uparrow^r}$]{$\tilde{x}_{n+1} = x_{n+1} {\uparrow^r}=\left(x_n{\downarrow_r}\right){\uparrow^r}$}, and replace it by its corresponding \emph{value} patch $V_j$ from the (sharp) source image at that scale, $x_n$. Key and value patches are trivially paired (have the same pixel coordinates). That way, blurry \emph{query} patches are compared with blurry \emph{key} patches, but are replaced with sharp \emph{value} patches.

Another difference regards the metric used to find nearest-neighbors. In some applications (e.g. image retargeting), it is essential to ensure that no visual data from the input is lost in the generated output. This was defined by \cite{simakov2008summarizing} as visual \emph{completeness}.
In \cite{simakov2008summarizing}, completeness is enforced using an iterative optimization process over the pixels. \mbox{InGAN} \cite{ingan} uses a cycle-consistency loss for this purpose (which comes at the expense of lost diversity). PNN enforces completeness using a \emph{normalized patch similarity score}, which replaces the common $L_2$-metric. This similarity score is used to find patch-nearest-neighbors, and favors \emph{key} patches that are not well-represented in the \emph{query}, practically encouraging visual completeness in an optimization-free manner, as detailed in step 3 of the algorithm summary below. \\



\begin{figure}
\centering
\includegraphics[width=\columnwidth]{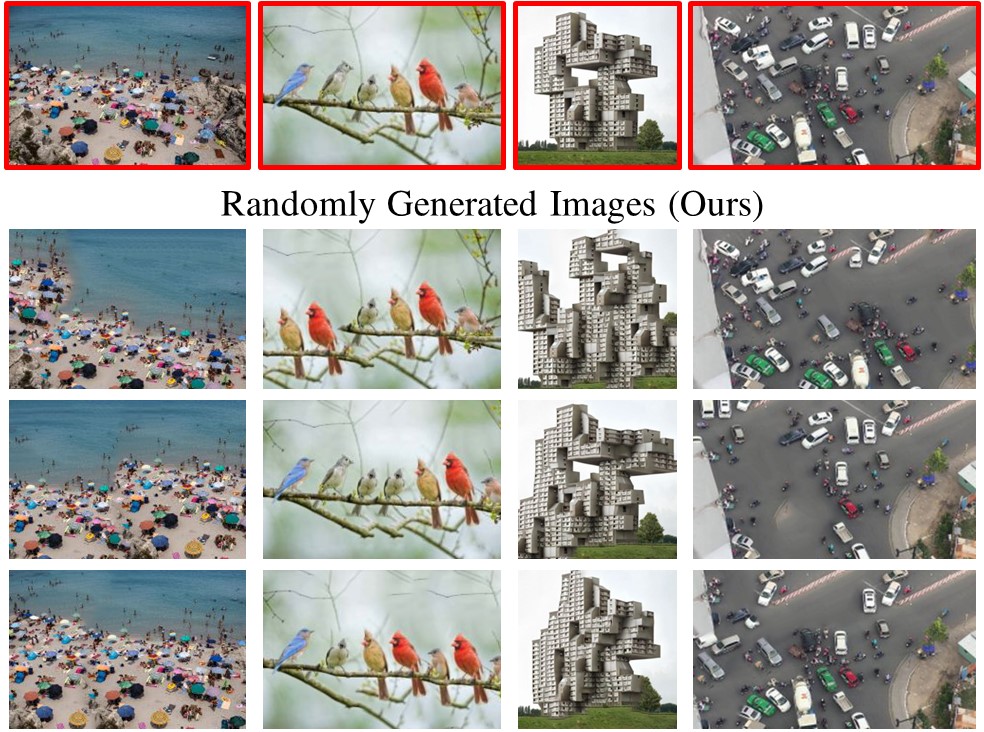}
\caption[]{\small \textbf{Diverse Image Generation:} \textit{(Please zoom in) Random images produced by GPNN from a single input (marked in red).}
    }
\label{fig:generation}
\vspace*{-0.7cm}
\end{figure}



\begin{table*}[h]
\vspace{-0.3cm}
\hspace{-0.15cm}
\begin{adjustbox}{max width=1.02\textwidth}
\begin{tabular}{clccccccccr}
\toprule

\multirow{2}{*}{\textbf{Dataset}} &
\multirow{2}{*}{{\textbf{Method}}} &
\textbf{SIFID} $\downarrow$  &
\textbf{NIQE} $\downarrow$  &
\multicolumn{2}{c}{\textbf{Confusion-Paired [$\%$]} $\uparrow$} &
\multicolumn{2}{c}{\textbf{Confusion-Unpaired [$\%$]} $\uparrow$} &
\textbf{Realism competition [$\%$]} $\uparrow$ &
\textbf{Diversity} &
\textbf{Runtime} $\downarrow$  \\

&&\cite{shaham2019singan}&\cite{mittal2012making}&Time Limit& No Time Limit &Time Limit& No Time Limit& GPNN vs. SinGAN &\cite{shaham2019singan}&[sec]
\\
\midrule

 & SinGAN ($N$)          & 0.085            & 5.240             &           21.5$\pm$1.5  & 22.5$\pm$2.5&           42.9$\pm$0.9  & 35.0$\pm$2.0 & 28.1$\pm$2.2 & 0.5  & 3888.0 \\ 
 & \textbf{GPNN} ($\sigma{=}1.25$)  & {\textbf{0.071}}   & {\textbf{5.049}}    &   {\textbf{44.7}}$\pm$1.7 & {\textbf{38.7}}$\pm$2.0 &   {\textbf{47.6}}$\pm$1.5 & {\textbf{45.8}}$\pm$1.6 & \textbf{71.9}$\pm$2.2 & 0.5 & \textbf{2.1} \\ \cmidrule{2-11}
 & SinGAN ($N{-}1$)        & 0.051            & 5.235             &           30.5$\pm$1.5  & 28.0$\pm$2.6 &    {\textbf{47}}$\pm$0.8  & 33.9$\pm$1.9 & 32.8$\pm$2.3 & 0.35 & 3888.0\\
 \multirow{-4}{*}{\shortstack{Places50 \\ \cite{zhou2016places, shaham2019singan}}}
 & \textbf{GPNN} ($\sigma{=}0.85$)  & {\textbf{0.044}}   & {\textbf{5.037}}    &   {\textbf{47}}$\pm$1.6   & {\textbf{42.6}}$\pm$1.7 &    {\textbf{47}}$\pm$1.4  & {\textbf{45.9}}$\pm$1.6 & \textbf{67.2}$\pm$2.3 & 0.35 & \textbf{2.1}\\ 
\midrule
 & SinGAN ($N$)          &        0.133     &   6.79            &              28.0$\pm$3.3              & 12.0$\pm$2.3 &   35.9$\pm$2.7                          & 39.9$\pm$2.7 & 41.7$\pm$3.3 &                 0.49 & 3888.0 \\ \multirow{-2}{*}{SIGD16}
 & \textbf{GPNN} ($\sigma{=}0.75$)  & {\textbf{0.07}}    & {\textbf{6.38}}     &   {\textbf{46.6}}$\pm$2.6                         & {\textbf{43.3}}$\pm$2.7 &    {\textbf{46.3}}$\pm$2.6                         & {\textbf{46.8}}$\pm$2.4 & \textbf{59.3}$\pm$3.3 &                0.52 & \textbf{2.1} \\ 
\bottomrule
\end{tabular}
\end{adjustbox}
\caption{\small \textbf{Quantitative Evaluation}. 
\textit{We evaluate our results over two datasets: The Place50 images used in the evaluation of~\cite{shaham2019singan}, and our new SIGD dataset (see text). We use a variety of measures: NIQE (unpaired image quality assessment)~\cite{mittal2012making}, SIFID - single image FID~\cite{shaham2019singan}, and human evaluations through an extensive user-study (see text). We repeat the evaluation for multiple diversity levels (measured as proposed by~\cite{shaham2019singan}). The table shows GPNN outperforms SinGAN by a large margin in every aspect: Visual quality, Realism, and Runtime.}
}
\label{table:comparison}
\vspace*{-0.5cm}

\end{table*}

\vspace{-0.3cm}
\noindent\mbox{\scalebox{0.95}[1]{\underline{PNN consists of 6 main algorithmic steps (numbered in Fig.~\ref{fig:pnn})}:}}
\begin{enumerate}[topsep=5pt,itemsep=-1ex,partopsep=1ex,parsep=1ex, leftmargin=*, wide=0pt]
    \item \textbf{Extract patches:} PNN receives a sharp source image $x_n$ and an initial guess $\tilde{y}_{n+1}$ (which is an upscaled version of the generated output from the previous scale, hence somewhat blurry). Patches from the initial guess are extracted into the \emph{query} pool of patches (denoted as $Q$). Nearest neighbors of the blurry query patches are searched in the similarly-blurry image obtained by upscaling the coarser source image, $\tilde{x}_{n+1}$ (Fig.~\ref{fig:method}(b)). Its patches are denoted as the \emph{key} pool of patches ($K$). The corresponding sharp patches are then extracted from the sharp source image $x_n$, and are denoted as the \emph{value} pool of patches ($V$). The only exception is the coarsest image scale ($n{=}N$), where we use the value patches as keys ($K=V$). The keys and values are ordered in the same way to maintain correspondences (patches from the same location have the same index in both pools). Patches are overlapping, so the same pixel can appear in multiple patches.
    
    \item \textbf{Compute Distances:} The MSE distance between each query patch $Q_i$ and each key patch $K_j$ is computed, and stored in the distances matrix $D_{i,j}$. \shai[Classical patch methods used this metric for nearest neighbor search. However, this one-directional metric on its own cannot ensure the \emph{completeness} term (for example, in Fig.~\ref{fig:pnn}(2), all queries are mapped to the same key, marked in red). In \cite{simakov2008summarizing} this was addressed by optimizing a bidirectional similarity measure. In PNN we solve this in an \emph{optimization-free} manner by using the \emph{normalized score} below.]{} We utilize the parallel nature of computing $D_{i,j}$, and run it on GPU for speed.
    \item \textbf{Compute Normalized Scores:} To encourage visual completeness, PNN uses a similarity score that favors key patches that are missing in the queries. This increases their chance to be chosen and appear in the output, and thereby improve \emph{Completeness}. The score normalizes the distance with a per-key factor:
    \vspace*{-0.2cm}
    \begin{equation}
    \label{eq:normalizedScore}
        S_{i,j} = \frac{D_{i,j}}{\alpha + \min_{\ell}{D_{\ell,j}}}
    \vspace*{-0.2cm}
    \end{equation}
    Intuitively, when a key patch $K_j$ is missing in the queries, the normalization term would be large and the score would be smaller. On the other hand, when a key patch appears in the queries, the normalization factor would get closer to $\alpha$. The parameter $\alpha$ is used as a knob to control the degree of completeness, where small $\alpha$ encourages completeness, and $\alpha \gg 1$ is essentially the same as using MSE. \shai[Note that this process requires no optimization (as opposed to \cite{simakov2008summarizing}).]{}

    \item \textbf{Find NNs:} For each query patch $Q_i$, we find the index of its closest \emph{key} patch, i.e. $j_{nn}(i) = \argmin_{\ell}{S_{i,\ell}}$.
    \item \textbf{Replace by NNs:} \shai[Using the Query-Key-Value scheme, w]{}We replace each query patch $Q_i$ with the value of its nearest neighbor, $V_{j_{nn}(i)}$. The output is denoted as $O_i$.
    \item \textbf{Combine Patches:} Overlapping patches are combined into an image. Pixels that appear in multiple overlapping patches are aggregated using a gaussian weighted-mean.
\end{enumerate}

Note that combining very different overlapping patches may cause inconsistencies and artifacts. To mitigate these inconsistencies, PNN is applied $T$ times at each scale (in our implementation $T{=}10$). In the first iteration, the initial guess is as explained above. In further iterations, the previous output (without upscaling) is used as initial guess. \ben[These iterations diffuse information from nearby patches, and help to obtain coherent final output.]{}

\vspace{0.3cm}
\textbf{Runtime:}
A key advantage of GPNN over GAN-based methods (e.g., SinGAN~\cite{shaham2019singan}, InGAN~\cite{ingan}, Structural Analogies~\cite{benaim2020structural}) is its short runtime. While GAN-based methods require a long training phase (hours \emph{per image}), GPNN uses a non-parametric generator which needs no training. Thus, SinGAN takes about $1$ \emph{hour} to generate a 180$\times$250 sized image, whereas GPNN does so in $2$ \emph{seconds} (see Table~\ref{table:comparison}).

\begin{figure*}
\vspace*{-0.3cm}
\centering
\includegraphics[width=\textwidth]{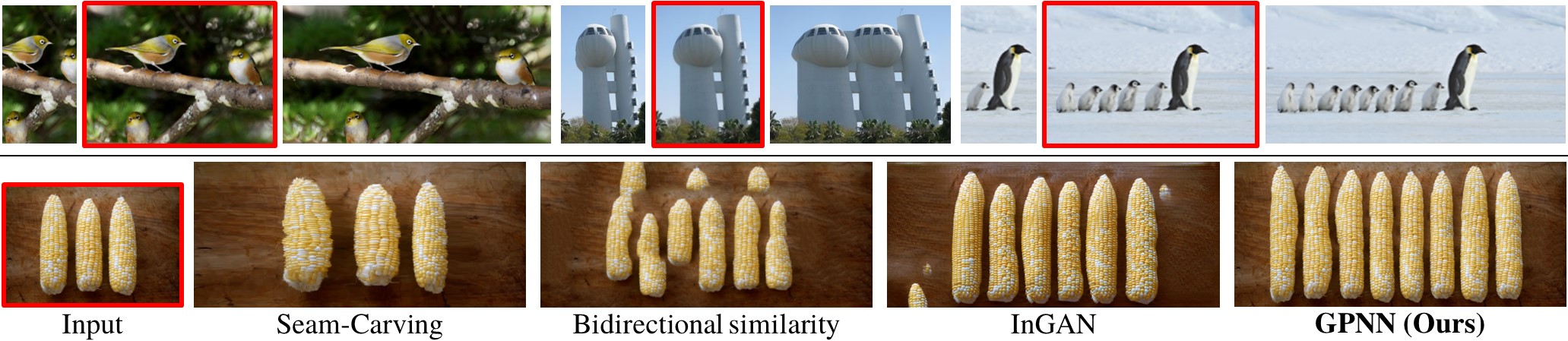}
\caption{\small \textbf{Retargeting:} \textit{(Please zoom in) Top rows show retargeted images by our method. Patch distribution is kept when retargeting to various target shapes. Bottom row shows comparison with previous patch-based \cite{avidan2007seam},\cite{simakov2008summarizing} and GAN-based \cite{ingan} methods.}}
\label{fig:retargeting}
\vspace*{-0.5cm}
\end{figure*}

\section{Results \& Evaluation}
\label{sec:results}
We evaluate and compare the performance of GPNN to SinGAN~\cite{shaham2019singan} on the main  application of \emph{random image generation}.
We first follow the exact same evaluation procedure of SinGAN \cite{shaham2019singan}, with the same data. We then add more measures, more tests and introduce more data. 
We show substantial supremacy in Visual-Quality and Realism with a large margin, both quantitatively and qualitatively, over all measures and datasets. The complete set of results can be found in the supplementary material. Runtime of GPNN is shown to be 3 orders of magnitude faster.



\noindent
\textbf{Data:} \ 
We evaluate our performance on 2 datasets. The first is the set of 50 images used in SinGAN~\cite{shaham2019singan} for their evaluation (50 images from Places365 dataset~\cite{zhou2016places}). The second is a new benchmark we introduce, Single Image Generation Dataset (SIGD) --  a set of 16 images that well exemplify a variety of different important aspects of single image generation tasks (visual aspects not represented in the structural Places images). These include:
7 images extracted from figures in the SinGAN paper~\cite{shaham2019singan}, 2 images from the Structural Analogies paper~\cite{benaim2020structural} and 7 more we collected from online sources. 
These SIGD images are characterized by having many more  conceivable versions per image than Places images, hence are more suited for comparing the quality of random image generation by different methods. 

\vspace{0.1cm}
\noindent
\textbf{Visual Results:} \ 
Fig.~\ref{fig:generation} shows GPNN results for diverse image generation, which highlight 3 characteristics of GPNN: (i)~\textit{Visual quality}: The results are sharp looking with almost no artifacts (please zoom-in). (ii)~\textit{Realistic structure}: The generated images look real, the structures make sense
(iii)~\textit{Diversity}: GPNN produces high diversity of results (e.g., diverse architectures of the building), while maintaining the above 2 characteristics. Fig.~\ref{fig:teaser}~(top-left) further shows results of GPNN on SinGAN's pivotal examples in their paper~\cite{shaham2019singan}.
%
Fig.~\ref{fig:generation_comparison} shows a visual comparison of GPNN to SinGAN.  Images generated by GPNN look very realistic, whereas SinGAN often produces artifacts/structures that make no sense (note that birds and branches generated by GPNN look very realistic despite the different ordering of the birds on the branch, while SinGAN doesn't maintain realistic structures).


\noindent
\textbf{Quantitative evaluation:} \ 
Table.~\ref{table:comparison} shows quantitative comparison between GPNN and SinGAN. 
All generated images are found in the supplementary material. We use the SIFID measure~\cite{shaham2019singan} to measure the distance between the source and the generated image patch distributions, as well as NIQE~\cite{mittal2012making} for reference-free quality evaluation. GPNN has much greater flexibility in choosing the degree of diversity (by tuning the input noise). However, for fair comparison,
we adjusted the input noise level in GPNN so that its diversity level matches that of SinGAN's results. Diversity is measured as proposed by SinGAN: pixelwise STD over 50 generated images. On SinGAN's places50 dataset we achieve clear superiority in both measures (SIFID \& NIQE), for both levels of diversity used in SinGAN. The margin is even larger on the SIGD dataset.

\vspace{0.1cm}
\noindent
\textbf{Qualitative Evaluation --  Extensive User-Study:} 
Table~\ref{table:comparison} displays the results of our user-study, conducted using Amazon Mechanical Turk platform.
\mbox{Our surveys composed of:} 
2 setups (paired \& unpaired) \textbf{$\times$} 2 datasets (Places50~\cite{shaham2019singan} \& SIGD) \textbf{$\times$} multiple diversity levels
\textbf{$\times$} 2 temporal modes (Time limit \& No time limit). 
Altogether, these resulted in \emph{27 different surveys, each answered by 50 human raters}.
The number of questions in each survey equals the number of images in the dataset. 
Results are summarized in Table~\ref{table:comparison}.

\noindent\emph{\underline{paired / unpaired setups}}:
In the paired setup, the ground-truth image and a generated image are shown side-by-side in random order. The rater is asked to determine which one is real. In the unpaired setup, a single image is shown (real or generated), and the rater has to decide whether it is real or fake. In both setups we report the percent of trials the rater was ``fooled" (hence, the highest expected score is 50\%).
The above setups were applied separately to GPNN and SinGAN. 
In addition, we also ran a \emph{Realism competition} where the rater has to decide which image looks more realistic, in  a paired survey of GPNN-vs-SinGAN (here the highest possible score is 100\%).

\noindent\emph{\underline{Time limit / No time limit}}:
We first followed the time-limited setup of SinGAN's user study~\cite{shaham2019singan}, which flashed each 
image for only 1 second (``Time limit").
We argue that $1sec$ makes it hard for raters to notice differences, resulting in a strong bias towards chance (50\%) for any method. We therefore repeat the study also with unrestricted time (``No time limit'').

\noindent\emph{\underline{Results}}:
GPNN scores significantly higher than SinGAN in all setups (Table~\ref{table:comparison}). 
Moreover, in the unlimited-time surveys (when the human rater had more time to observe the images), SinGAN's ability to ``fool'' the rater significantly drops, especially in the unpaired case. In contrast, having no time-limit had a very small effect on GPNN's confusion rate. 
\michal{Looking at the table, I don't think we can claim these last 2 sentences.}
In all surveys GPNN got results very close to chance level ($50\%$). The fact that an observer with unlimited time can rarely distinguish between real and GPNN generated images, implies high realism of the generated results.
Finally, in the direct  GPNN-vs-SinGAN survey (unlimited time), GPNN's results were selected as more realistic than SinGAN's for most images in all surveys.


\begin{figure}
    \includegraphics[width=\columnwidth]{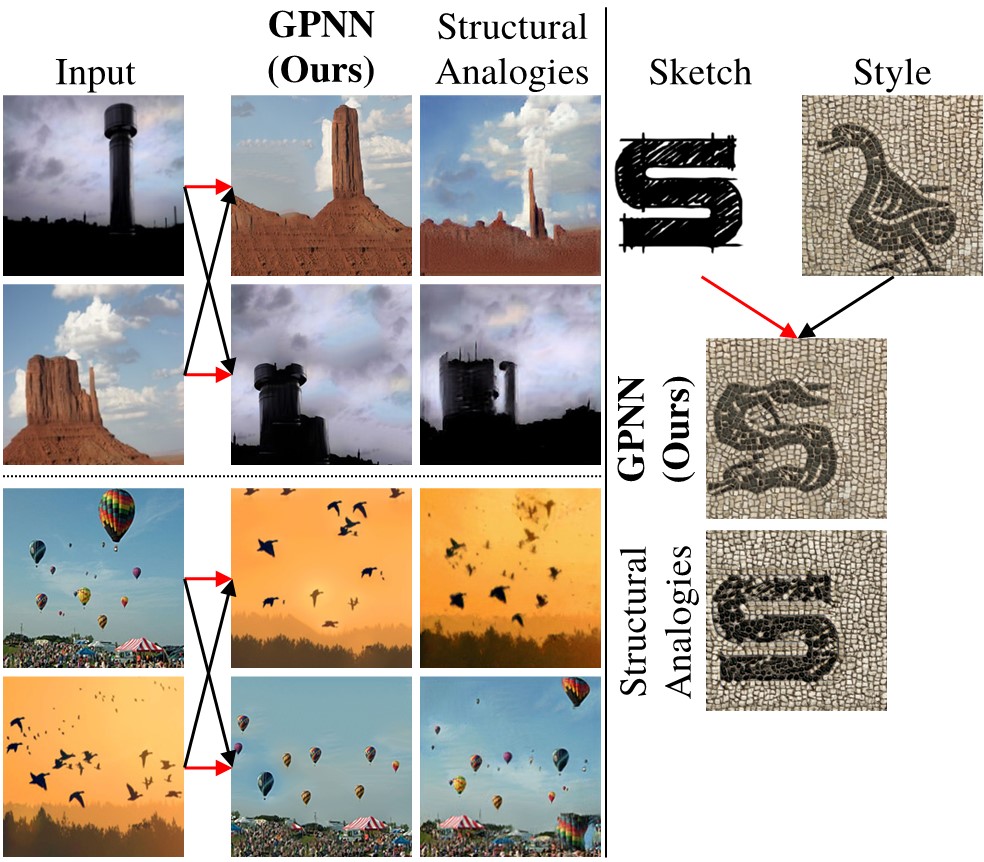}
    \caption{\small \textbf{Structural Analogies:} \textit{(Please zoom in) 
    Red arrow indicates the `structure', while black arrow indicates the `source' (defining the patch distribution to match). GPNN is compared to~\cite{benaim2020structural}. GPNN can also generate sketch-to-image instances (right).
    }}
    \label{fig:structural_analogies}
    \vspace*{-0.6cm}
\end{figure}

\section{Additional Applications}
\label{sec:apps}
In addition to diverse image generation, GPNN gives rise to 
many other applications (old and new), all within a \emph{single unified framework}. The different applications are obtained simply by modifying a few basic parameters in GPNN, such as the pyramid depth $N$, the initial guess at the coarsest level $\tilde{y}_{N+1}$, and the choice of the hyper-parameter $\alpha$ in Eq.~\ref{eq:normalizedScore}.
We next describe each application, along with its design choices.

\vspace*{0.1cm}
\noindent
\textbf{Retargeting:} 
The goal is to resize the single source image to a target size (smaller or larger; possibly of different aspect ratio), but \emph{maintain the patch distribution of the source image} (i.e., maintain the size, shape and aspect-ratio of all small elements of the source image) \cite{ingan, simakov2008summarizing, barnes2009patchmatch}. GPNN starts by naively resizing the input image to the target size, and downscale it by a factor of $r^N$, this is inserted as the initial guess $\tilde{y}_{N+1}$. The pyramid depth $N$ might change according to the objects size in the image, but it's usually chosen such that the smaller axis of the image at the coarsest pyramid level is roughly $\times 4$ GPNN's patch size $p$. For retargeting, we wish to retain as much visual information as possible from the source image, and hence $\alpha$ in Eq.~\ref{eq:normalizedScore} is set to a small value (e.g., $\alpha=.005$), thus promoting ``Completeness''. To get better final results, the described process is done gradually (similarly to \cite{simakov2008summarizing}), i.e. resizing in a few small steps.
Results for retargeting can be seen in Fig.~\ref{fig:retargeting} and \ref{fig:teaser}. Fig.~\ref{fig:retargeting} further compares the performance of our method with that of \cite{ingan, simakov2008summarizing, avidan2007seam}. GPNN produces results which are more realistic, and with less artifacts. 


\vspace*{0.1cm}
\noindent
\textbf{Image-to-image and Structural Analogies:} We demonstrate image-to-image translations of several types. There exist many approaches and various goals and brandings for image-to-image translations; Style-transfer, domain-transfer, structural-analogies \cite{ulyanov2017improved, gatys2016image, kolkin2019style, zhu2017unpaired, johnson2016perceptual, isola2017image, liao2017visual, Elad_2017}.
Given two input images $\mathnormal{A}$ and $\mathnormal{B}$, we wish to create an image with the patch distribution of $\mathnormal{A}$, but which is structurally aligned with $\mathnormal{B}$. Namely, a new image where all objects are located in the same locations as in $\mathnormal{B}$, but with the visual content of $\mathnormal{A}$. For that, GPNN sets the source image $x$ to be $\mathnormal{A}$. Our initial guess $\tilde{y}_{N+1}$, is chosen as $\mathnormal{B}$ downscaled by $r^N$. This guess sets the overall structure and location of objects, while GPNN ensure that the output has similar patch distribution to $A$. The pyramid depth $N$ may change between pairs of images according to the change in object sizes. The output should contain as much visual data from $\mathnormal{A}$ (e.g., in the bottom-left pair in Fig.~\ref{fig:structural_analogies}, it is desired that many types of balloons will appear in the output), hence we set $\alpha$ in Eq.~\ref{eq:normalizedScore} to be small (e.g., $\alpha=.005$). Finally, for refinement of the output, the output is downscaled by $r^N$ and re-inserted to GPNN. Results can be found in Fig.  \ref{fig:structural_analogies}. Our method creates new objects at the locations of objects of image $\mathnormal{B}$, while the output image seems to be from the distribution of image $\mathnormal{A}$ as desired. GPNN works well also for sketch-to-image instances as shown. Compared to the GAN-based method of \cite{benaim2020structural}, our results suffer from less artifacts, and are more reliable to the style of the source image $\mathnormal{A}$ (e.g., the ``S'' image in Fig. \ref{fig:structural_analogies}). In addition to providing superior results, GPNN is also several orders of magnitude faster than \cite{benaim2020structural}.

\vspace*{0.1cm}
\noindent
\textbf{Conditional Inpainting:} Similarly to the well studied inpainting task, in this task an input image with some occluded part is received, and the missing part should be reconstructed. However, in our suggested conditional version, in addition to regular image completion, the user can further steer the way the missing part is filled. This is obtained by the user marking the image region to be completed, with a region of \emph{uniform color of choice}, which is the "steering direction" (e.g., blue to fill sky, green to guide the completion toward grass, etc.). This will be the source image $x$. Note that GPNN gets the mask of the occluded part, hence does not use patches which originated from there. The initial guess $\tilde{y}_{N+1}$ is set to be a downscaled version of $x$ by $r^N$. The number of pyramid levels $N$ is chosen such that the occluded part in the downscaled image is roughly $p \times p$ (the size of a single patch). PNN applied at the coarsest level replaces the masked part coherently with respect to the chosen color. In finer levels, details and textures are added. In this task, completeness is not required, hence a large $\alpha$ is set in Eq.~\ref{eq:normalizedScore}. Fig.~\ref{fig:teaser}, show the choice of different colors for the same masked region indeed affects the outcome, while maintaining coherent and visually appealing images.

\vspace*{0.1cm}
\noindent
\textbf{Image Collage:} 
This task, previously demonstrated in~\cite{simakov2008summarizing}, aims
to \emph{seamlessly} merge a set of $n$ input images $\{{x^i}\}_{i=1}^n$ to a single output image, so that no information/patch from the input images is lost in the output (``Completeness''), yet, no new visual artifacts that did not exist in the inputs are introduced in the output (``Coherence''). We create the initial guess $\tilde{y}_{N+1}$ by first naively concatenating the input images. Then, we use the same design as for retargeting, with a single change - GPNN extracts patches from all the source images (rather than from a single source image in retargeting). 
Fig.~\ref{fig:collage} shows a collage produced by GPNN, 
on an example taken from~\cite{simakov2008summarizing}.
Compared with~\cite{simakov2008summarizing}, our results are sharper and more faithful to the inputs.

\vspace*{0.1cm}
\noindent
\textbf{Image Editing:} 
In image editing/reshuffling~\cite{simakov2008summarizing, shaham2019singan}, 
one makes a change in the image (move objects, add new objects, change locations, etc.), and the goal is to seamlessly blend the change in the output image. We use the unedited original image as the source image $x$, and a downscaled version of the edited image by $r^N$ as the initial guess $\tilde{y}_{N+1}$. Completeness is not required in this task (e.g., if the edit removes an object), thus we set $\alpha$ in Eq.~\ref{eq:normalizedScore} to be large.
The depth $N$ of pyramid is set such that the edited region covers roughly the size of a single patch ($p \times p$) in the coarsest scale. Similarly to inpainting, the area around and inside the edited region is "corrected" by our algorithm to achieve coherence. Noise may be added at the coarsest level,  to allow for different coherent solutions given a single input. Fig.~\ref{fig:editing} shows our editing results compared to SinGAN's \cite{shaham2019singan}. Our results tend to be less blurry (especially visible around the edited region).





\narrowstyle
\section{\mbox{{GANs vs. Patch Nearest-Neighbors: Pros \& Cons}}}
\label{sec:discussion}
\normalstyle

The experiments in Secs.~\ref{sec:results} and~\ref{sec:apps} show striking superiority of GPNN compared to single-image GANs, both in visual-quality and in run-time (while having comparable diversity).
This section first analyzes the source of these surprising \emph{inherent advantages} of simple classical patch-based methods (exemplified through GPNN).
%
%
Nevertheless, single-image GANs have several significant capabilities which \emph{cannot be realized} by simple patch nearest-neighbor methods. Hence, despite the title of our paper, you may not always want to ``Drop the GAN''... These \emph{inherent limitations} of classical patch-based methods (GPNN included) are also discussed.

%


\subsection*{\underline{Advantages}}
\label{sec:advantages}
\vspace{-0.2cm}
\noindent
The advantages of Patch-based methods over GANs stem primarily from one basic fundamental difference: Single-image GANs \emph{implicitly learn} the  patch-distribution  of  a  single  image, whereas classical Patch-based approaches \emph{explicitly maintain} the entire patch-distribution  (the image itself),  and directly access it via  patch  nearest-neighbor  search.  This fundamental difference yields the following advantages:

\begin{figure}[t]
\vspace*{-0.1cm}
    \centering
\includegraphics[width=\columnwidth]{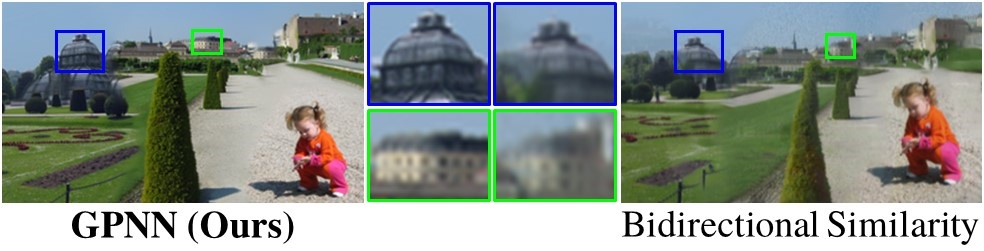}
\caption{\small \textbf{Collage:} 
\textit{Multiple input images are seamlessly combined into a single coherent output, maintaining visual information from all inputs. Note the higher quality of GPNN compared to Bidiectional-Similarity~\cite{simakov2008summarizing}.
\textcolor{red}{The 3 input images  are found  in Fig.\ref{fig:teaser}.}}
}
\label{fig:collage}
\vspace*{-0.5cm}
\end{figure}

\begin{figure}[t]
\vspace*{-0.5cm}
    \centering
    \includegraphics[width=\columnwidth]{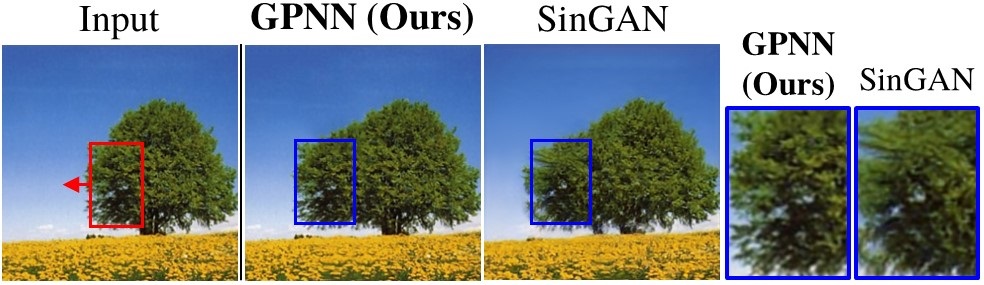}
    \caption{\small \textbf{Image Editing:} \textit{A naively edited image is injected to our pyramid of PNNs. Compared with the results of \cite{shaham2019singan}.}}
    \label{fig:editing}
\vspace*{-0.5cm}
\end{figure}

\vspace*{0.1cm}
\noindent\textbf{Visual Quality:}
An output image produced by patch nearest-neighbor search, is \emph{composed of original image patches} pulled out directly from the input image. In contrast, in GANs the output is \emph{synthesized via an optimization process}. GPNN thus produces images whose patches are more faithful to the original input patches than GANs.
This yields sharper outputs (almost as sharp as the input), 
with fewer undesired visual artifacts (please zoom in on Fig.~\ref{fig:generation_comparison}, \ref{fig:editing} to compare).

\noindent\textbf{Runtime:}
Since no training takes place,  the runtime of patch-based methods reduces \emph{from hours to seconds} compared to GANs (Table~\ref{table:comparison}). Moreover, since nearest-neighbor search can be done independently and in parallel for different image patches, this naturally leverages GPU computing.

\noindent\textbf{Visual Completeness:}  While GANs are trained to produce patches of high likelihood (thus encouraging output \textit{Coherence} to some extent), no mechanism enforces \textit{Completeness} (i.e., encourage all patches of the input image to appear in the output image). 
Lack of Completeness is further intensified by the natural tendency of GANs to suffer from mode collapse. 
In contrast, classical patch-based methods can explicitly enforce Completeness, e.g., by optimizing \emph{Bidirectional patch similarity} between the input and output image~\cite{simakov2008summarizing}, or in an optimization-free manner by using GPNN's \emph{patch-specific} normalized score (Eq.~(\ref{eq:normalizedScore})).
InGAN~\cite{ingan} has successfully introduced Completeness into GANs, by training a \emph{conditional} single-image GAN via an  encoder-encoder scheme with a reconstruction loss.
However, this came with a  price: lack of diversity in the generated outputs.
In contrast, GPNN is able to promote both Completeness \& Coherence, 
as well as large output diversity. 
There is an  \emph{inherent trade-off between the Completeness and Diversity}. The $\alpha$ parameter in Eq.~(\ref{eq:normalizedScore})  thus
provides a ``knob'' to control the degree of desired Completeness in GPNN
(according to the image/application at hand).
Despite this new flexibility of GPNN compared to GANs, all experiments in Table~\ref{table:comparison} were performed with a fixed $\alpha$ (for fairness).

\noindent\textbf{Visual Coherence (realistic image structures):}
The iterative nearest-neighbor search of classical patch-based methods prevents forming adjacency of output patches that are not found adjacent in the input image (for any image scale). This tends to generate realistic looking structures in the output. In contrast, such tendency for coherence is only weakly enforced in GANs. Proximity between unrelated patches may emerge in the generated output, since the generator is fully convolutional with limited receptive field. The generator typically generates mitigating pixels, hopefully with high likelihood. This often results in incoherent non-realistic image structures and artifacts that do not exist in the input image.
See examples in Fig.~\ref{fig:generation_comparison} and in Supplementary-Material.

\noindent\textbf{Controlling diversity vs. global structure:} There is a natural trade-off between high output diversity and preserving global image structure. The magnitude $\sigma$ of the noise added in GPNN to the coarsest scale of the input image, provides a simple user-friendly ``knob'' to control the degree of desired output diversity. It further yields a natural \emph{continuum} between large diversity (high noise) and  global structural fidelity (low noise).
GANs, on the other hand, do not hold any mechanism for controlling the preservation of global structure (although there are some inductive biases that tend to preserve global structure implicitly~\cite{xu2020positional}).
While GANs may support a few discrete levels of diversity (e.g.,~\cite{shaham2019singan} demonstrates 2 diversity levels), this diversity is not adjustable.

\subsection*{\underline{Limitations}}
\label{sec:limitations}
\vspace{-0.2cm}
\noindent\textbf{Generalization:}
Classical patch-based methods use a \emph{discrete} patches distribution. GANs on the other hand learn a \emph{continuous} distribution. GANs can therefore generate novel patches with high likelihood from the learned distribution. This  capability is lacking in patch-based methods. Such generalization can be advantageous (e.g., image harmonization~\cite{shaham2019singan}), but may also be disadvantageous, as it frequently generates undesired artifacts (see above).

\begin{figure}
\vspace*{-0.5cm}
    \centering
\includegraphics[width=0.95\columnwidth]{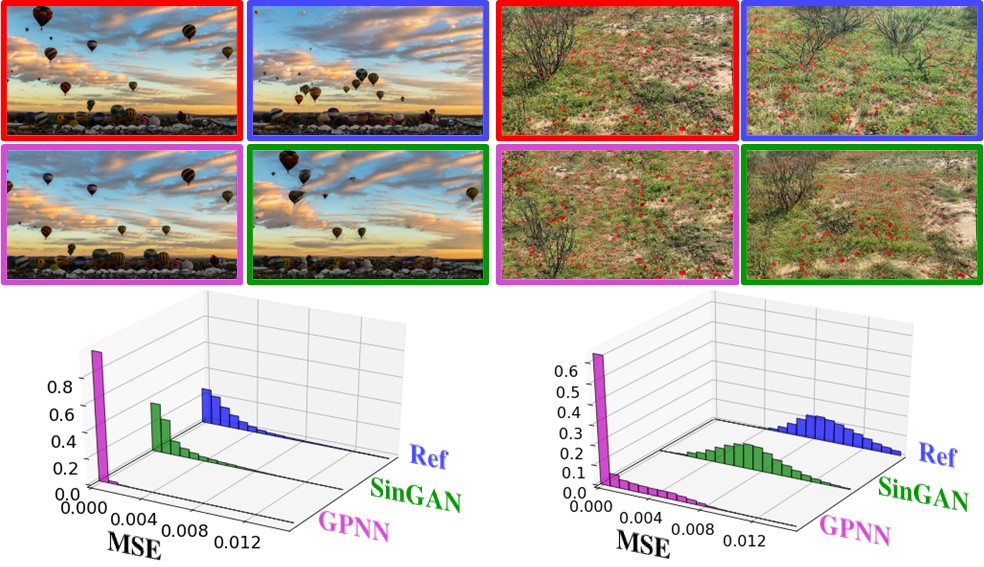} 
\caption{\small  \textbf{Are GANs ``Fancy Nearest Neighbors Extractors"?} 
{\it Input image (\textcolor{in_red}{red}), reference (\textcolor{ref_blue}{blue}), GPNN generated (\textcolor{gpnn_purple}{purple}), SinGAN generated (\textcolor{gan_green}{green}). Please see text for details.}
}
\label{fig:discussion}
\vspace*{-0.5cm}
\end{figure}

\noindent\textbf{Continuous output generation:}
Neural networks are continuous functions. Small change in the latent input causes a small divergence in the generated output. This enables latent space interpolation and other smooth manipulations, such as single image animation~\cite{shaham2019singan}, or smooth resizing animation~\cite{ingan}. 
In contrast, nearest neighbor search is discrete in nature.  This prevents naively performing continuous interpolation or animation in classical patch-based methods.

\noindent\textbf{Mapping to patches vs. Mapping to pixels:}
In classical nearest-neighbor methods (including GPNN), the nearest-neighbor search maximizes the quality of the extracted patches, but not the quality of the final output pixels. The formation of the output image typically involves heuristic averaging of overlapping patches. This may introduce some local blurriness in patch-based methods.
GAN discriminators also judge {output patches} in the size of their receptive field. However, since
the generators receive pixel-based gradients, they can \emph{optimize directly for each output pixel}. 
\assaf{Originally I mentioned SR in this context. I think we should since one possible question a reviewer may ask is what about SR since demonstrated in SinGAN. I think we would be doing ourselves a big favor if we mention the apps we cannot do in the eubmission rather than in the rebuttal.}

\subsection*{Are GANs ``Fancy Nearest Neighbors Extractors"?}
\noindent
It has been argued that GANs are only elaborated machinery for nearest neighbors retrieval.
In \emph{single-image} GANs, the ``dataset" is small enough (a single image), providing an excellent  opportunity to quantitatively examine this claim. 
Fig.~\ref{fig:discussion} shows two generated random instances of the same input image (red), once using SinGAN (green) and once using GPNN (purple).
We also added a reference image (blue) of the same scene taken 
a slightly different position/time
representing an ``ideally" new generated instance.
We measured the distance between the generated patches to their nearest neighbors in the  input image and plotted the  histograms of these distances (MSE).
We repeated this experiment twice for 2 different input images.
It is easy to see that the GAN-generated patch distribution behaves more like the reference distribution than the patch nearest neighbor generated distribution.
That is, GANs are capable of generating new samples beyond nearest neighbors.
However, this capability of GANs comes with a price, its newly generated instances suffer from blur and artifacts (\emph{please zoom in}).
In contrast, GPNN generated samples has higher fidelity to the input (the histogram is peaked at zero), resulting with higher output quality, but at a limited ability to generate novel patches. 

\section{Acknowledgements}
This project received funding from the European Research Council (ERC) under the European Union’s Horizon 2020 research and innovation programme (grant agreement No 788535). Dr Bagon is a Robin Chemers Neustein AI Fellow.

\clearpage
{\small
\bibliographystyle{ieee_fullname}
\bibliography{main}
}
\end{document}